\documentclass[sigconf,nonacm]{acmart}
\AtBeginDocument{%
  }

\usepackage{graphicx}
\usepackage{amsmath}
\usepackage{amsthm}
\usepackage{booktabs}
\usepackage{algorithm}
\usepackage{algorithmic}
\usepackage[switch]{lineno}
\usepackage{bbding}
\usepackage{multirow}
\usepackage{subfig}
\usepackage[misc]{ifsym}
\begin{document}

\title{Efficient Agent: Optimizing Planning Capability for Multimodal Retrieval Augmented Generation}


\author{Yuechen Wang\textsuperscript{1},  Yuming Qiao\textsuperscript{1}, Dan Meng\textsuperscript{1}\textsuperscript{\Letter} \and Jun Yang\textsuperscript{2}, Haonan Lu\textsuperscript{2}, Zhenyu Yang\textsuperscript{2}, Xudong Zhang\textsuperscript{1}}
\affiliation{
\institution{\textsuperscript{1}OPPO Research Institute}
\institution{\textsuperscript{2}OPPO AI Center}
\country{\textsuperscript{\Letter}mengdan90@163.com}
}







\renewcommand{\shortauthors}{Wang et al.}

\begin{abstract}
Multimodal Retrieval-Augmented Generation (mRAG) has emerged as a promising solution to address the temporal limitations of Multimodal Large Language Models (MLLMs) in real-world scenarios like news analysis and trending topics. However, existing approaches often suffer from rigid retrieval strategies and under-utilization of visual information.
To bridge this gap, we propose E-Agent, an agent framework featuring two key innovations: a mRAG planner trained to dynamically orchestrate multimodal tools based on contextual reasoning, and a task executor employing tool-aware execution sequencing to implement optimized mRAG workflows.
E-Agent adopts a one-time mRAG planning strategy that enables efficient information retrieval while minimizing redundant tool invocations.
To rigorously assess the planning capabilities of mRAG systems, we introduce the Real-World mRAG Planning (RemPlan) benchmark. This novel benchmark contains both retrieval-dependent and retrieval-independent question types, systematically annotated with essential retrieval tools required for each instance. 
The benchmark’s explicit mRAG planning annotations and diverse question design enhance its practical relevance by simulating real-world scenarios requiring dynamic mRAG decisions.
Experiments across RemPlan and three established benchmarks demonstrate E-Agent’s superiority: 13\% accuracy gain over state-of-the-art mRAG methods while reducing redundant searches by 37\%.
\end{abstract}





\maketitle

\section{Introduction}

\begin{figure}
    \centering
    \includegraphics[width=0.99\linewidth]{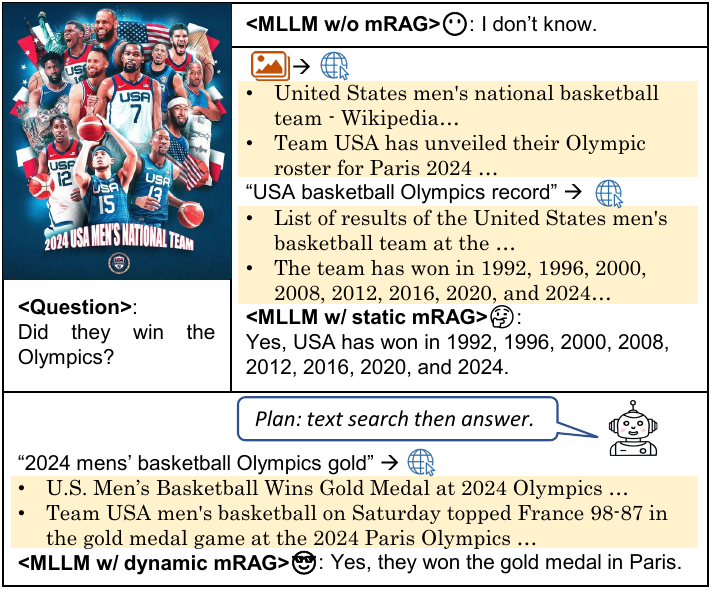}
    \caption{Comparison among VQA systems without RAG, with static mRAG, and with dynamic mRAG.}
    \label{fig:fig1}
\end{figure}

The burgeoning field of Visual Question Answering (VQA) has witnessed growing interest in enhancing system capabilities through Retrieval-Augmented Generation (RAG), particularly for handling complex queries requiring external knowledge. While current VQA systems demonstrate proficiency in straightforward tasks, they exhibit notable deficiencies when confronted with questions demanding extensive domain knowledge or timely information. These limitations underscore the critical need for methodological innovations that can bridge the gap between conventional approaches and real-world application requirements.

Recent advancements in Multimodal Retrieval-Augmented Generation (mRAG) systems have sought to augment Large Language Models (LLMs) by integrating internet search capabilities for accessing specialized knowledge.
Early practices typically employ a two-stage process: Multimodal Large Language Models (MLLMs) first generate visual captions, followed by text-based retrieval through LLMs~\cite{lin-byrne-2022-retrieval}. However, such methods predominantly rely on textual information processing, failing to fully exploit multimodal data sources. This text-centric paradigm significantly constrains system effectiveness, particularly when handling image-based queries or multimodal information needs~\cite{zhang2024vision}. For instance, conventional search engines remain fundamentally incapable of processing visual content directly, resulting in incomplete information retrieval.

Emerging research attempts to address these limitations through multimodal retrieval tools, yet current implementations maintain rigid, predetermined workflows~\cite{Hu2023RevealRV,Jiang2024MMSearchBT}. 
These static architectures lack the adaptive capability to dynamically select appropriate search modalities based on query characteristics, leading to suboptimal knowledge retrieval and compromised answer quality. 
The limitations become particularly pronounced in scenarios requiring real-time information updates or cross-domain reasoning, where inflexible retrieval strategies often yield redundant searches and irrelevant results. This not only degrades system efficiency but also introduces noise that adversely impacts response accuracy.

Recent work by OmniSearch~\cite{li2024benchmarking} proposes an adaptive planning framework that decomposes complex queries into subproblems for multimodal retrieval. While demonstrating improved flexibility through real-time feedback mechanisms, its iterative planning approach incurs significant computational overhead and latency. This stepwise decision-making process frequently leads to inefficient resource utilization and incomplete execution paths, ultimately undermining the practical viability of mRAG systems.


Motivated by these identified limitations in current multimodal retrieval paradigms, we propose Efficient Agent (E-Agent), a novel agent framework that performs multimodal input comprehension, single-pass mRAG planning, and optimized execution of search and MLLM operations.
Our framework eliminates redundant search operations through deterministic planning while maintaining adaptability through dynamic tool selection.
By decoupling planning from execution, E-Agent significantly reduces error propagation risks inherent in feedback-dependent systems. 
Notably, the architecture operates effectively with an 8B parameter model, substantially lowering computational requirements compared to existing planning approaches.

To establish rigorous evaluation standards for this emerging research direction, we introduce the Real-World mRAG Planning (RemPlan) benchmark, the first comprehensive testbed specifically designed for assessing dynamic multimodal retrieval planning capabilities. 
RemPlan features diverse question types and image sources, making it closer to real-world applications.
Each piece of collated data in RemPlan is meticulously annotated with standard mRAG plan alongside the corresponding answers. 
Furthermore, we develop a hierarchical plan evaluation metric that elevates evaluation beyond conventional answer accuracy measurements. This novel assessment protocol calculates mRAG planning accuracy, search tool precision \& recall, and parameter semantic scores. 
We conduct extensive comparison experiments on RemPlan and other mRAG datasets. Experimental results validate the effectiveness of E-Agent, and demonstrate the superiority of the new mRAG benchmark RemPlan.


In summary, our contributions are threefold:
\begin{itemize}
    \item \textbf{E-Agent Framework}: A novel plan-then-execute architecture combining a dynamic mRAG planner with a tool-aware executor, achieving state-of-the-art performance in VQA tasks through optimized multimodal retrieval workflows.
    \item \textbf{RemPlan Benchmark}: The first comprehensive evaluation framework for mRAG systems, featuring retrieval-dependent/-independent questions with expert-validated plans and disentangled evaluation protocols.
    \item \textbf{Empirical Validation}: Extensive experiments demonstrating improvements of VQA ability improvements alongside considerable reduction in redundant searches, supported by systematic analysis studies.
\end{itemize}
We anticipate that our work will advance the development of intelligent multimodal QA systems through its methodological innovations and rigorous evaluation framework. 

\begin{figure*}[tbh]
    \centering
    \includegraphics[width=0.99\linewidth]{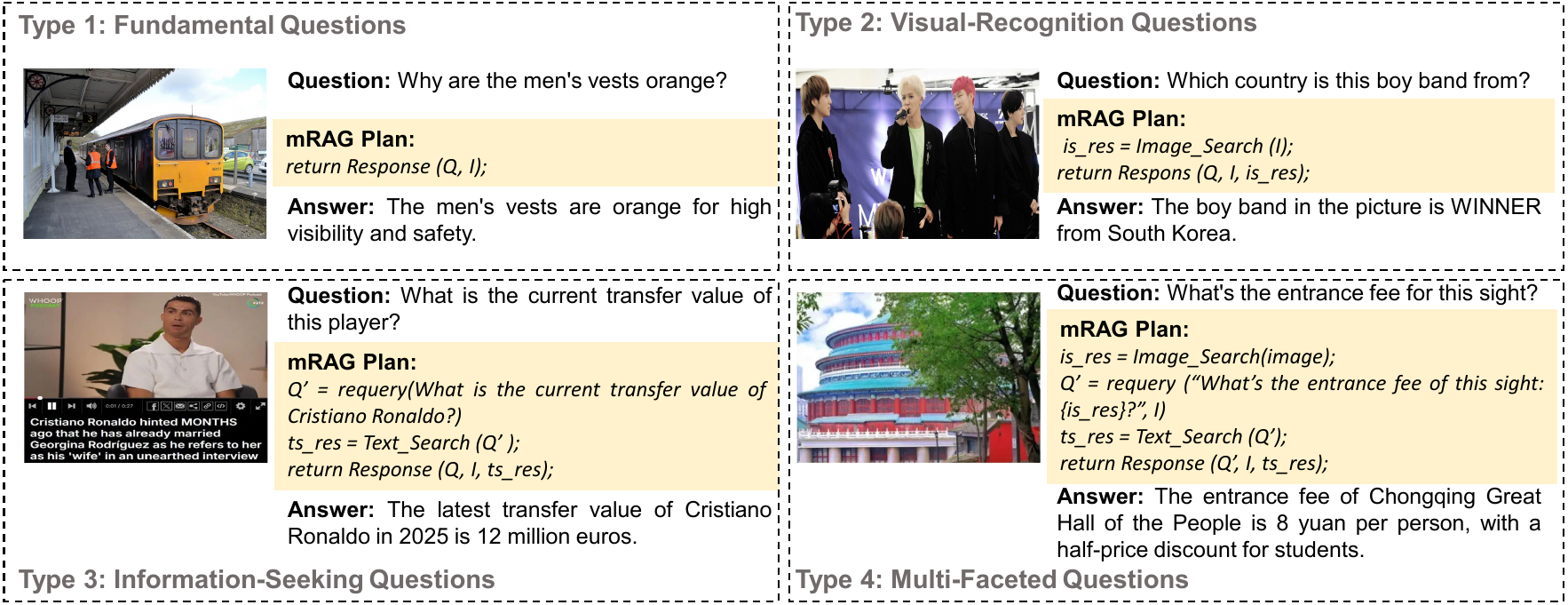}
    \caption{Data samples of different question types in RemPlan.}
    \label{fig:question_types}
\end{figure*}

\section{Related Work}
\subsection{Multimodal Retrieval Augmented Generation}
Retrieval Augmented Generation (RAG) has established itself as an effective paradigm for enhancing language models with external knowledge while maintaining reasoning capabilities~\cite{borgeaud2022improving,NEURIPS2020_6b493230,10516270,ram2023context}.
Recently, the emergence of Multimodal Large Language Models (MLLMs)~\cite{alayrac2022flamingo,liu2024llavanext}has extended this paradigm to multimodal contexts, with mRAG demonstrating promising applications~\cite{zhao-etal-2023-retrieving}.

Early mRAG approaches primarily focused on visual feature extraction through standalone vision models combined with text-based retrieval~\cite{lin2022retrieval,yang2022empirical,lin2023fine}.
Subsequent work leveraged image search engines for visual similarity matching, utilizing retrieved web content to assist visual question answering~\cite{Jiang2024MMSearchBT}.
However, these methods adopted fixed retrieval pipelines that often introduced computational overhead and irrelevant information due to their static architecture.
Recent advancements like Vision Search Assistant~\cite{zhang2024vision} and OmniSearch~\cite{li2024benchmarking} introduced dynamic tool integration during reasoning processes. While improving flexibility, these approaches suffer from redundant execution pipelines and repetitive reasoning steps that compromise system efficiency. 
This reveals a critical gap in existing mRAG systems’ ability to dynamically coordinate retrieval operations with intrinsic model capabilities - a core innovation the proposed E-Agent achieves through its efficient contextual planning mechanism.


\subsection{Knowledge-Based Visual Question Answering Benchmarks}
The evolution of Visual Question Answering (VQA) benchmarks has progressively emphasized knowledge-intensive reasoning since its inception~\cite{doi:10.1073/pnas.1422953112}.
Early datasets like KBQA~\cite{10.5555/3171642.3171825} and FVQA~\cite{8046084} focused on structured knowledge graphs, while OK-VQA~\cite{Marino_2019_CVPR} and its extensions~\cite{10.1145/3404835.3463259,Schwenk2022AOKVQAAB} shifted toward open-domain commonsense reasoning. 
Subsequent benchmarks including KVQA~\cite{10.1609/aaai.v33i01.33018876}, ViQuAE~\cite{10.1145/3477495.3531753} and INFOSEEK~\cite{chen-etal-2023-pre-trained} required external knowledge retrieval for accurate responses.
However, the knowledge required by these datasets may be readily absorbed by large-scale pretrained models through standard training procedures.

Recently, several benchmarks are proposed for mRAG evaluation~\cite{Jiang2024MMSearchBT, li2024benchmarking}, which contain fact-asking questions that require searching for newest information or specialized knowledge to answer.
While these benchmarks assess basic retrieval capabilities through answer verification, their evaluation frameworks exhibit three critical limitations: (1) Over-reliance on search result quality and MLLM capacities, (2) Inability to measure advanced planning, tool orchestration, and reasoning skills essential for agent-based mRAG systems, and (3) Universal assumption of mandatory external retrieval for all questions, which may encourage unnecessary retrieval operations in practical deployments.

To overcome these limitations, we propose RemPlan, the first benchmark featuring dynamic mRAG planning, explicit tool-use annotations, and diverse questions requiring dynamic mRAG decisions.

\section{The RemPlan Benchmark}

In this section, we present the Real-World mRAG Planning (RemPlan) benchmark, a novel evaluation framework designed to systematically assess mRAG capabilities in VQA systems. 
This benchmark specifically addresses the critical need for evaluating dynamic planning strategies in real-world multimodal reasoning scenarios.

\subsection{Dataset Construction}
\label{sec:data_construction}
The RemPlan dataset was developed through a four-stage construction process combining manual and automated approaches:

\textbf{Image Collection.}
Our multimodal corpus integrates two primary image sources: real-world VQA data from application scenarios and news-related imagery from diverse web resources. 
The collected images underwent a rigorous quality control pipeline involving automated deduplication followed by expert manual review to eliminate low-resolution or irrelevant visual content. This curation process ensures dataset diversity while maintaining high visual quality standards.

\textbf{Question Annotation.}
Each image undergoes further processing, during which human annotators are tasked with writing questions about the provided image. These annotators are also required to tag each question with notes indicating whether visual recognition or the retrieval of external information is necessary for an accurate answer. The question annotation process ensures the accuracy of plans generated in the subsequent stages.

\textbf{Plan Generation.}
Leveraging GPT-4o’s advanced reasoning capabilities, we generated formal mRAG execution plans based on annotated image-question pairs. 
Each plan specifies two key components: a formatted multimodal tool invocation sequence, as well as the tool argument values. 

\textbf{Human Verification and Answer Annotation.}
A panel of domain experts conducted final validation by checking the plan feasibility, and evaluating consistency between questions and required mRAG operations. They are also required to annotate the question answers based on web-sourced information. 
The expert panel comprises postgraduate degree holders with certified advanced English proficiency, who possess a comprehensive understanding of both the search tools and MLLMs.
This multi-stage verification ensures both semantic validity and practical relevance of the benchmark instances.

Upon completion of these steps, the Real-World mRAG Planning (RemPlan) dataset stands with a robust collection of 200 Image-Question pairs, each annotated with mRAG planning trajectories and answers. This comprehensive process has established a high-quality dataset, paving the way for an effective and realistic evaluation of mRAG systems.

\subsection{Dataset Analysis}
\label{dataset_analysis}

\begin{table*}[tbh]
\centering
\caption{Comparison of RemPlan and other Information-seeking VQA datasets.}
\begin{tabular}{lccccc}
\toprule
Features & A-OKVQA & InfoSeek & MMSearch & Dyn-VQA & \textbf{RemPlan} \\
\midrule
\multicolumn{6}{c}{\textit{Question Types}} \\
\midrule
Fundamental            & \Checkmark & \Checkmark & \XSolidBrush & \XSolidBrush & \Checkmark \\
Visual-Recognition     & \XSolidBrush & \XSolidBrush & \XSolidBrush & \Checkmark & \Checkmark \\
Information-Seeking    & \XSolidBrush & \Checkmark & \Checkmark & \Checkmark & \Checkmark \\
Multi-Faceted          & \XSolidBrush & \XSolidBrush & \Checkmark & \Checkmark & \Checkmark \\
\midrule
\multicolumn{6}{c}{\textit{Annotations}} \\
\midrule
Required mRAG Tools  & \XSolidBrush & \XSolidBrush & \XSolidBrush & \XSolidBrush & \Checkmark \\
\bottomrule
\end{tabular}
\label{tab:data_compare}
\end{table*}

The RemPlan benchmark introduces several key enhancements and distinguishing characteristics that address critical gaps in existing evaluation frameworks, as detailed in the following subsections.

\subsubsection{Diversity of Questions, Images, and Answers}

The questions involved in real-world multimodal question answering can be divided into 4 categories with regard to the required search type:
\begin{itemize}
\item \textbf{Type 1: Fundamental Questions.} These questions can be addressed utilizing pretrained knowledge, thereby eliminating any need for additional search tools.
\item \textbf{Type 2: Visual-Recognition Questions.} This category includes questions necessitating image search in order to identify specific visual elements, such as distinguishing certain people, organisms or locations.
\item \textbf{Type 3: Information-Seeking Questions.} Representing a step-up in complexity, these questions call for a comprehensive web-based research to capture up-to-date or specialized knowledge not typically included in pretraining corpus.
\item \textbf{Type 4: Multi-Faceted Questions.} In this most complex classification, questions demand both visual recognition and external information retrieval, thus requiring both image search and text search to answer.
\end{itemize}
In figure~\ref{fig:question_types}, we show examples of each type of question.

\begin{figure}[tbh]
    \centering
    \subfloat[Question distribution]{
    \includegraphics[width=0.8\linewidth]{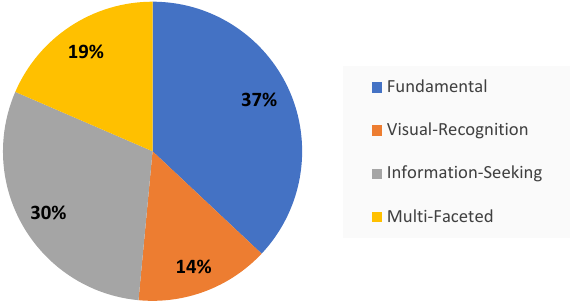}
    \label{fig:question_distribution}
    } \\
    \subfloat[Image diversity and answer length in mRAG datasets]{
    \includegraphics[width=0.9\linewidth]{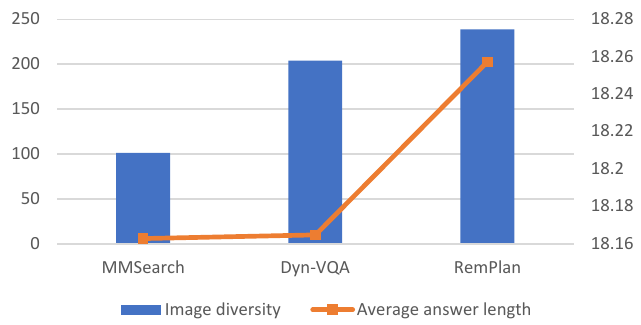}
    \label{fig:image_diversity_and_answer_length}
    }
    \caption{Statistics of RemPlan dataset.}
\end{figure}

One prominent characteristic of RemPlan is its incorporation of all four types of questions within the above taxonomy. This allows for a detailed assessment of agents' abilities in discerning whether any search is necessary, and if so, what type of search is required. 
The proportions of the four types of questions in RemPlan are illustrated in Figure~\ref{fig:question_distribution}.
As shown in Table~\ref{tab:data_compare}, most of the existing benchmarks, including traditional VQA datasets and datasets designed for mRAG of MLLMs, encompass only a subset of these four question types. In contrast, RemPlan features a more diverse and balanced distribution of questions, rendering it an excellent platform for evaluating mRAG methods across various scenarios. 

Beyond question taxonomy, RemPlan advances dataset realism through two key dimensions. As shown in Figure~\ref{fig:image_diversity_and_answer_length}, RemPlan has increased image diversity\footnote{The image diversity score is calculated by finding the Shannon entropy based on the image similarity matrix following~\cite{friedman2022vendi}.} and longer answer length in RemPlan compared to other datasets. 
This characteristic enables a more authentic evaluation of the agents' performance in real-world settings. 

\subsubsection{Disentangled mRAG Evaluation Protocol}
RemPlan pioneers a novel annotation scheme that decouples planning evaluation from tool execution outcomes.
As illustrated in the last section, each instance in RemPlan includes both final answer and mRAG planning trajectory annotations, making it possible to assess the performance of mRAG planning module by studying the accuracy of searching tool usage, independent from the confounding effects of downstream tool performance. 
By leveraging the direct planning evaluation, researchers can gain more granular insights into how well the models employ various multimodal search tools and devise effective retrieval strategies.

\begin{figure*}[tbh]
    \centering
    \includegraphics[width=0.9\linewidth]{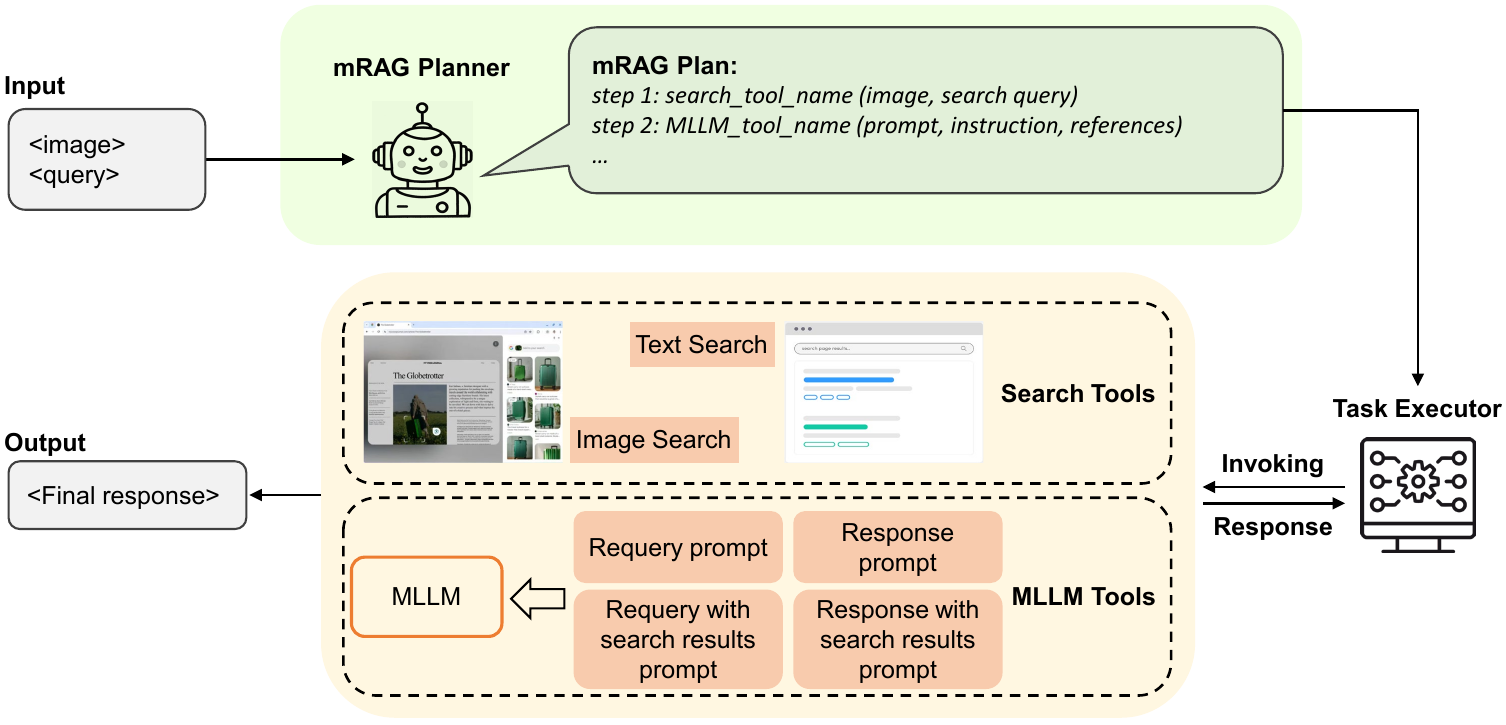}
    \caption{The E-Agent framework}
    \label{fig:framework}
\end{figure*}

\subsection{Plan Evaluation Metrics}
\label{section: Plan Evaluation Metrics}
The unique capability of the proposed Real-World mRAG Planning (RemPlan) Benchmark to directly evaluate agents' planning abilities necessitates a rigorous evaluation method. 
Accordingly, we introduce a comprehensive set of metrics specifically designed to evaluate mRAG planning trajectories effectively.

\begin{itemize}
\item \textbf{Tool-Specific Precision and Recall.} These metrics evaluate each search tool's precision and recall in all the planning result, which reflect the ability of agent to understand and invoke different search tools. 
The precision and recall of images search is noted as `IS-P', `IS-R', and the precision and recall of text search is noted as `TS-P', `TS-R'.
\item \textbf{Plan Accuracy.} Goes beyond the evaluation of invoking search tools, this metric evaluates whether the agent can correctly arrange the MLLM to cooperate with search tools and provides a complete and correct plan. This metric is noted as `Plan-acc'.
\item \textbf{Parameter Correctness.} This metric appraises the validity of the parameters used for invoking the search tools and MLLM, providing insights into the agent's ability to manipulate tools effectively. This metric is noted as `Param-acc'.
\item \textbf{Semantic Similarity.} This metric includes an evaluation of the semantic consistency between the natural language parameters utilized in the planning process and the annotated ground truth. By doing so, it aims to assess if (1) the planning retains the semantic intent of the original user question, and (2) the query used during the search operation appropriately reflects the knowledge needed to be retrieved, thereby ensuring semantic consistency and valid knowledge retrieval.
This metric is noted as `Param-sim'.
\end{itemize}

The above evaluation method lends a new perspective to the assessment process, ensuring a comprehensive examination of the mRAG agents’ planning abilities.

\section{The E-Agent Framework}

In this paper, we propose a novel framework aiming to optimize planning for multimodal retrieval-augmented generation, named Efficient Agent (E-Agent). 
As depicted in Figure~\ref{fig:framework}, the E-Agent framework operates through two interconnected modules: the \textbf{mRAG planner} and the \textbf{Task Executor}. The mRAG planner determines the sequence of actions, deciding when to employ search tools and when to rely on the MLLMs directly. The \textbf{Task Executor} then carries out these actions, either by leveraging MLLMs or by combining search tools with MLLMs as necessary.

\subsection{mRAG planner}
In contrast to conventional static mRAG systems that employ fixed execution pipelines regardless of query context, the mRAG planner in E-Agent performs contextual analysis of both textual queries and visual inputs through a single forward pass to formulate a comprehensive mRAG plan. 
This unified planning strategy simultaneously determines three critical components: 
(1) optimal selection of multimodal search tools based on needed information, 
(2) adaptive configuration of auxiliary MLLM function, 
and (3) generation of specialized instructions and parameters for various tool invocation.

By the dynamic mRAG planning mechanism, the E-Agent framework can retrieve more precise external knowledge through context-aware tool selection.
Moreover, this one-time planning approach eliminates redundant search iterations while maintaining computational efficiency - contrary to conventional multi-stage decision-making pipelines that often accumulate multiple inference overheads.


\begin{table*}[tbh]
\centering
\caption{Performance comparison among mRAG methods on RemPlan benchmark. Type 1-4 refers to question types introduced in Section~\ref{dataset_analysis}}
\begin{tabular*}{0.95\linewidth}{@{}lcccccccccccc@{}}
\toprule
\multirow{2}{*}{Method} & \multicolumn{5}{c}{Answer quality (Ans.)}  & \multicolumn{7}{c}{Plan evaluation metric} \\ 
\cmidrule(r){2-6} \cmidrule{7-13} 
 & Type 1 & Type 2 & Type 3 & Type 4 & All & IS-P & IS-R & TS-P & TS-R & Plan-acc & Param-acc & Param-sim \\
\midrule
Qwen2-VL-72B & 1.54 & 0.72 & 0.88 & 0.84 & 1.09 & \textcolor{gray}{0.00} & \textcolor{gray}{0.00} & \textcolor{gray}{0.00} & \textcolor{gray}{0.00} & - & - & - \\
\midrule
MMSearch & 0.49 & 0.93 & 0.53 & 0.41 & 0.55 & 0.33 & \textbf{1.00} & 0.49 & \textbf{1.00} & - & - & - \\
OmniSearch & 0.90 & 0.62 & 0.80 & 0.84 & 0.8
2 & 0.57 & 0.85 & 0.60 & 0.92 & - & - & - \\
\midrule
\textbf{E-Agent}-fewshot & 1.53 & \textbf{1.24} & \textbf{1.04} & \textbf{0.93} & 1.23 & 0.46 & 0.95 & 0.76 & 0.61 & 0.32 & 0.71 & 0.91 \\
\textbf{E-Agent}-sft & \textbf{1.65} & 1.17 & 1.00 & 0.89 & \textbf{1.25} & \textbf{0.85} & 0.93 & \textbf{0.93} & 0.92 & \textbf{0.86} & \textbf{0.96} & \textbf{0.94} \\
\bottomrule
\end{tabular*}
\label{tab:res_remplan}
\end{table*}

\begin{table}[t]
\centering
\caption{Number of tool calls of different methods on the RemPlan benchmark.}
\resizebox{0.99\linewidth}{!}{
\begin{tabular}{lccc}
\toprule 
Method & Search Tools & MLLM & mRAG planner \\ 
\midrule
Qwen2-VL-72B & \textcolor{gray}{0.00} & \textcolor{gray}{1.00} & \textcolor{gray}{0.00} \\
\midrule
MMSearch & 2.00 & 3.00 & \textbf{0.00} \\
OmniSearch & 1.96 & 1.96 & 2.96 \\
\midrule
\textbf{E-Agent}-fewshot & \textbf{1.05} & 1.77 & 1.00 \\
\textbf{E-Agent}-sft & \textbf{1.05} & \textbf{1.54} & 1.00 \\
\bottomrule
\end{tabular}
}
\label{tab:res_remplan_efficiency}
\end{table}

\subsection{Task Executor}
The Task Executor serves as the implementation engine that translates the structured plan into executable actions. This component invokes designated search tools and MLLMs according to parameter specifications in the generated plan. 
Furthermore, it dynamically selects context-appropriate prompt templates for the MLLM tools in the mRAG plan.

Within the this pipeline, the MLLM serves as various functions depending on the trajectory generated by the mRAG planner.
We implement the MLLM tools using the Qwen2-VL-72B model, with manually written task-specific prompt templates.
In addition, to incorporate essential external knowledge with both visual and textual queries, the system employs dual-modality search interfaces.
The MLLM tools and search tools are list as follows:

\textbf{(1) Requery tool.} 
This MLLM-driven component synthesizes visual inputs (original image and possible image search results) and textual queries to formulate optimized search strings for subsequent text retrieval.
The generated search query employ concise phrase structures emphasizing key informational elements, diverging from neutral language question.

\textbf{(2) Response tool.} Operating as the terminal processing unit, this MLLM component aggregates the input image, query, potential image search, and text search results to produce coherent, user-oriented responses.

\textbf{(3) Image search tool.} Our image search API connects to web-scale reverse image search services, returning relevant webpage content through similarity-based visual matching.
This capability enables cross-modal identification of entities that are not explicitly encoded in MLLM parameters, such as public figures, flora/fauna species, and geographic locations.

\textbf{(4) Text search tool.} 
This search engine integration executes keyword-based web queries using compact text phrases, accessing real-time information updates and domain-specific knowledge beyond the MLLM’s pretraining corpus scope.

All tool configurations receive explicit representation during the training process of mRAG planner. 


\section{Experiments}
\subsection{Datasets and Metrics}
To comprehensively evaluate the performance of the proposed method, we conducted experiments on the RemPlan benchmark as well as three related datasets: the MMSearch benchmark~\cite{Jiang2024MMSearchBT}, Dyn-VQA~\cite{li2024benchmarking}, and A-OKVQA~\cite{Schwenk2022AOKVQAAB}.
As illustrated in Table~\ref{tab:data_compare}, the MMSearch benchmark and the Dyn-VQA dataset are designed specifically for mRAG, in which all the questions require multimodal retrieval to answer.
A-OKVQA is a traditional information-seeking VQA dataset presented in a multiple-choice format. 

On RemPlan benchmark, except for the plan evaluation metrics defined in Section~\ref{section: Plan Evaluation Metrics}, showing the planning ability of mRAG method directly,
we also report the final answer score evaluated by GPT-4o, denoted as `Ans.'.
In the process of evaluation, GPT-4o is prompted to gauge a score within the range of 0 to 2, contingent on the corresponding image, query, ground-truth answer, and the model's response. A score of 0 signifies a totally incorrect response, whereas a score of 2 denotes an entirely accurate answer.
The Ans. score is also reported for Dyn-VQA and MMSearch dataset.
For the A-OKVQA dataset, given its multiple-choice format, we employ answer accuracy as the evaluation metric.
Furthermore, we report the average number of calls of search tools, MLLM, and mRAG planner on all datasets, which indicates the average cost of different methods to answer questions.

\subsection{Experimental Settings}

In the experiments, we select Qwen2-VL-72B~\cite{Qwen2VL} as the MLLM backbone. We utilize \textbf{Baidu Image Search}~\footnote{https://image.baidu.com/} as the image search tool and \textbf{Tavily}~\footnote{https://tavily.com/} as the text search engine.
For the mRAG planner, we employ InternVL2-8B~\cite{chen2024expanding}, fine-tuning it with a training set comprising 10K data samples. The training data contains images, questions, and plan annotations. The collection of training data is the same as in Section~\ref{sec:data_construction} while the human verification and answer annotation phase are excluded.

For fair comparison of the performance among the E-Agent and other mRAG methods, we have reproduced MMSearch~\cite{Jiang2024MMSearchBT} and OmniSearch~\cite{li2024benchmarking} using the same MLLM backbone and search tools as in E-Agent. Additionally, we report the results of the raw MLLM without any searching to offer a more comprehensive analysis.

\subsection{Results on the RemPlan benchmark}
In Table~\ref{tab:res_remplan}, we compare the performance of E-Agent and other methods on the RemPlan benchmark. The overall question-answering quality is evaluated by GPT-4o and reported for each question type and for the whole benchmark. We also report the plan evaluation results based on the mRAG plan annotation in RemPlan, which reflects the precision and efficiency of search tool usage in different methods.

As shown in the results, the proposed E-Agent achieves state-of-the-art VQA performance on all four types of questions in the RemPlan benchmark. 
Specifically, E-Agent outperforms other mRAG methods by a large margin for Type 1 questions, which require no searching. In fact, MMSearch and OmniSearch perform even worse than the base MLLM, due to the noise brought by redundant searching.
E-Agent also achieves much better answer scores for Type 2 and Type 3 questions, owing to the more precise search tool usage.
This underscores the importance of determining when searching is necessary in more generalized application scenarios, where a considerable part of questions can be answered based on MLLM's inherent knowledge. 

Table~\ref{tab:res_remplan_efficiency} shows a comparative analysis about the number of tool calls. These statistics demonstrate that the one-time dynamic planning strategy employed by E-Agent remarkably diminishes the tool invocation frequencies, both for search tools and auxiliary MLLMs. 
This reduction is evident when compared to the static mRAG method (MMSearch) and the dynamic mRAG method (OmniSearch). Consequently, E-Agent enhances both the performance and efficiency of question-answering (QA) systems.

To validate the reliability of GPT-4o evaluation, which is reported as the Ans. metric for answer quality evaluation, we measure the consistency of the Ans. metric and human evaluation.
Specifically, we ask human evaluators to score answers of different methods on the RemPlan benchmark and calculate the Pearson correlation between Ans. score and human evaluation score.
As shown in Table~\ref{tab:score_correlation}, the Ans. metric shows high correlation with human evaluation, which illustrates the reliability of the Ans. metric.

\begin{table}[t]
\centering
\caption{Consistency between the Ans. score and human evaluation score.}
\begin{tabular}{lccc}
\toprule
Method & Ans. & Human eval. & Correlation \\ 
\midrule
Qwen2-VL-72B & 1.09 & 1.11 & 0.72 \\
MMSearch & 0.55 & 0.90 & 0.69 \\
OmniSearch & 0.82 & 1.02 & 0.74 \\
E-Agent & 1.25 & 1.42 & 0.78 \\
\bottomrule
\end{tabular}
\label{tab:score_correlation}
\end{table}

\begin{table}[t]
\centering
\caption{Results on Dyn-VQA benchmark.}
\resizebox{0.99\linewidth}{!}{
\begin{tabular}{lcccc}
\toprule
\multirow{2}{*}{Method} & \multirow{2}{*}{Ans.} & \multicolumn{3}{c}{Average Number of Calls} \\ \cmidrule{3-5} 
 &  & Search Tools & MLLM & mRAG planner \\ 
\midrule
Qwen2-VL-72B & 0.60 & \textcolor{gray}{0.00} & \textcolor{gray}{1.00} & \textcolor{gray}{0.00} \\
\midrule
MMSearch & 0.81 & 2.00 & 3.00 & \textbf{0.00} \\
OmniSearch & 0.82 & 2.03 & 2.03 & 3.03 \\
\midrule
\textbf{E-Agent}-fewshot & 0.77 & 1.74 & \textbf{1.86} & 1.00 \\
\textbf{E-Agent}-sft & \textbf{0.89} & \textbf{1.26} & 1.91 & 1.00 \\
\bottomrule
\end{tabular}
}
\label{tab:res_dynvqa}
\end{table}

\begin{table}[t]
\centering
\caption{Results on MMSearch benchmark.}
\resizebox{0.99\linewidth}{!}{
\begin{tabular}{lcccc}
\toprule
\multirow{2}{*}{Method} & \multirow{2}{*}{Ans.} & \multicolumn{3}{c}{Average Number of Calls} \\ \cmidrule{3-5} 
 &  & Search Tools & MLLM & mRAG planner \\ 
\midrule
Qwen2-VL-72B & 0.41 & \textcolor{gray}{0.00} & \textcolor{gray}{1.00} & \textcolor{gray}{0.00} \\
\midrule
MMSearch & 0.65 & 2.00 & 3.00 & \textbf{0.00} \\
OmniSearch & 0.58 & 2.35 & 2.35 & 3.35 \\
\midrule
\textbf{E-Agent}-fewshot & 0.64 & 1.77 & \textbf{1.71} & 1.00 \\
\textbf{E-Agent}-sft & \textbf{0.76} & \textbf{1.42} & 1.85 & 1.00 \\
\bottomrule
\end{tabular}
}
\label{tab:res_mmsearch}
\end{table}

\subsection{Results on other mRAG benchmarks}
The experimental results on Dyn-VQA and the MMSearch benchmark are shown in Table~\ref{tab:res_dynvqa} and ~\ref{tab:res_mmsearch}.

Compared to the above results on the RemPlan benchmark, these two benchmarks require at least one time of searching for all test samples. 
The superior performance of E-Agent on these benchmarks shows the efficacy of the one-time dynamic mRAG planning.
Even when searching is known to be required, a pre-generated plan of necessary searching tools still helps to improve answer quality and remove redundant searching brought by uncertainty.

Moreover, it can be concluded from the last two rows in Table~\ref{tab:res_remplan}, ~\ref{tab:res_dynvqa}, and~\ref{tab:res_mmsearch} that the supervised fine-tuned version of E-Agent achieves more efficient mRAG and generates better answers on both in-domain (RemPlan) and out-of-distribution (Dyn-VQA and MMSearch) benchmarks.
This observation proves that the trained E-Agent has a good generalization ability for more accurate mRAG planning.

\subsection{Results on Traditional VQA datasets}
In addition to the benchmarks specifically designed for mRAG, we also evaluated the methods on A-OKVQA. As shown in Table~\ref{tab:res_aokvqa}, the performance of E-Agent was comparable to the baseline, whereas other mRAG methods underperformed relative to the baseline.

Through a detailed analysis of the results, we observed that the errors made by the baseline model predominantly pertain to issues related to visual understanding or inaccuracies within the ground truth, rather than questions requiring external information. 
On the one hand, this observation substantiates that our approach effectively mitigates the performance and efficiency drawbacks associated with excessive searching in QA systems.
On the other hand, the experimental results further underscore the inadequacy of traditional VQA datasets in gauging a model’s capability to acquire external knowledge in the era of large models. Such datasets can be considered to encapsulate information that large models already possess as `common knowledge', further emphasizing the significance of introducing new mRAG benchmarks.

\subsection{Case Analysis}

\begin{figure*}[tbh]
    \centering
    \includegraphics[width=0.99\linewidth]{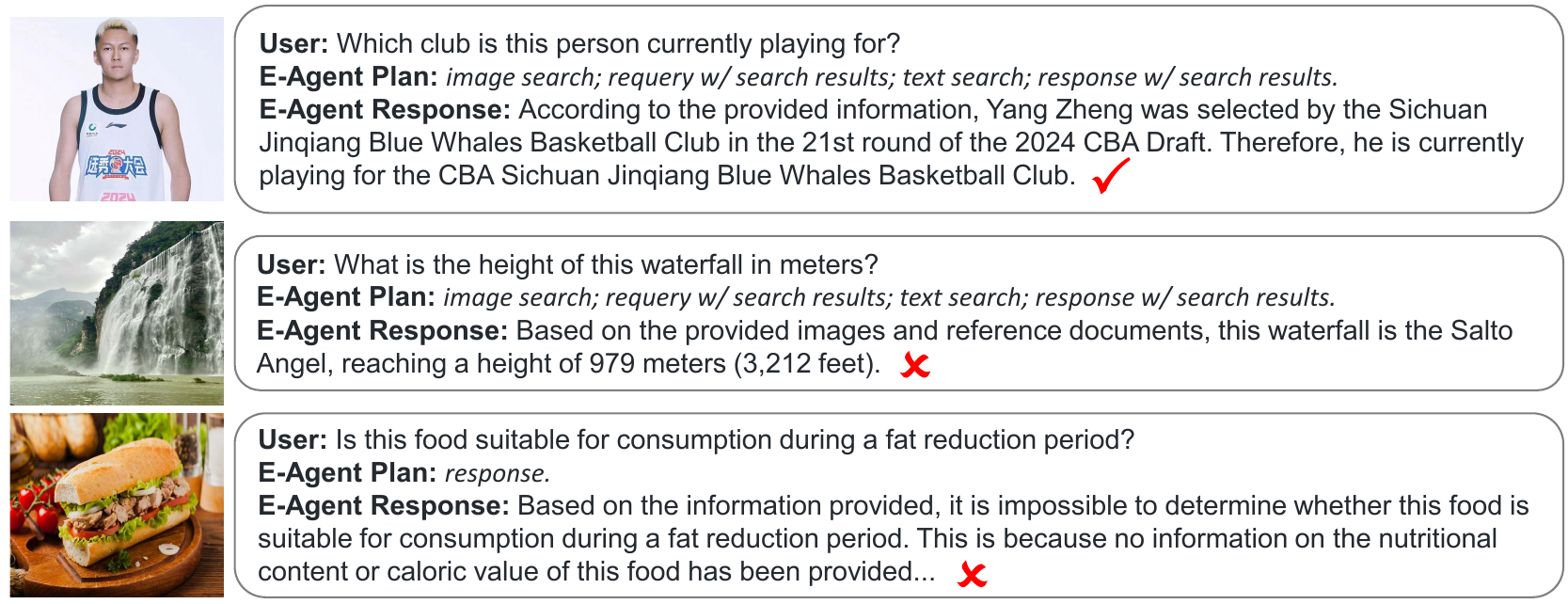}
    \caption{Case visualization of E-Agent predictions on the RemPlan benchmark. E-Agent successfully planned to perform both image search and text search in the first case, and obtained correct answer. While in the second and third case, either the planning or the searching tool failed, resulting in incorrect answer.}
    \label{fig:case_visualization}
\end{figure*}

\begin{table}[t]
\centering
\caption{Results on A-OKVQA benchmark.}
\resizebox{0.99\linewidth}{!}{
\begin{tabular}{lcccc}
\toprule
\multirow{2}{*}{Method} & \multirow{2}{*}{Acc} & \multicolumn{3}{c}{Average Number of Calls} \\ \cmidrule{3-5} 
 &  & Search Tools & MLLM & mRAG planner \\ 
\midrule
Qwen2-VL-72B & \textbf{0.88} & \textcolor{gray}{0.00} & \textcolor{gray}{1.00} & \textcolor{gray}{0.00} \\
\midrule
MMSearch & 0.67 & 2.00 & 3.00 & \textbf{0.00} \\
OmniSearch & 0.79 & 1.26 & 1.26 & 1.26 \\
\midrule
\textbf{E-Agent}-fewshot & 0.87 & 1.02 & 1.45 & 1.00 \\
\textbf{E-Agent}-sft & \textbf{0.88} & \textbf{0.13} & \textbf{1.02} & 1.00 \\
\bottomrule
\end{tabular}
}
\label{tab:res_aokvqa}
\end{table}

Figure~\ref{fig:case_visualization} illustrates several representative cases from the RemPlan benchmark along with the corresponding mRAG plan and final response generated by E-Agent, demonstrating both successful and suboptimal planning scenarios.

The first case exemplifies a successful scenario where the mRAG planner formulates the correct plan: (1) use image search tool to recognize the person in the picture, (2) use the requery tool by feeding a requery prompt to the MLLM, generating the search query based on the image search results, (3) use text search tool to obtain the latest information, and (4) use the response tool by feeding a response prompt to MLLM to summarize the search results and generate the final answer.

The following two cases are less successful. 
In the second case, although the mRAG planner successfully generates a reasonable mRAG plan, which first invokes the image search tool to recognize the identity of the waterfall, followed by searching for information about it and generating a response. 
However, the image search result is completely incorrect, leading to a wrong answer. This case indicates that the quality of the search results significantly impacts the VQA performance.
In the third case, the mRAG planner failed to plan necessary searching to recognize the food or retrieve the nutritional information, leaving the problem to the base MLLM. Therefore, the MLLM can only give an uncertain answer. These contrasting cases collectively demonstrate the dual challenges of effective tool orchestration and retrieval reliability in real-world mRAG applications.

\section{Conclusion and Discussion}
This study establishes a novel framework for advancing multimodal information retrieval through three key innovations: (1) systematic analysis of current mRAG limitations, (2) development of an adaptive planning architecture, and (3) creation of a scientifically constructed evaluation benchmark. Our proposed E-Agent framework introduces a paradigm shift in mRAG implementation by decoupling strategic planning from operational execution, enabling simultaneous improvements in accuracy and efficiency.

The architecture's single-pass planning mechanism demonstrates particular effectiveness in real-world VQA scenarios, achieving 13\% higher accuracy than state-of-the-art methods while reducing redundant searches by 37\%. Furthermore, the accompanying RemPlan benchmark addresses a critical gap in mRAG evaluation through its systematically annotated dataset containing diverse expert-validated image-question-plan-answer tuples and novel multidimensional assessment metrics.



While the proposed plan-then-execute agent framework demonstrates robust performance in general scenarios, two principal limitations merit discussion. First, the current implementation faces challenges when handling complex multi-hop reasoning tasks requiring iterative plan refinement, as its one-shot planning mechanism lacks intermediate verification steps. Second, the framework's dependence on predefined toolkits necessitates periodic updates to maintain compatibility with evolving multimodal data sources, potentially limiting long-term adaptability. These limitations suggest the following promising research directions.
First, we might develop hierarchical planning architectures that combine high-level strategy formulation with fine-grained plan adjustment through intermediate validation checkpoints.
Furthermore, dynamic reflection modules can be integrated that are capable of real-time plan optimization based on retrieval feedback loops.
On top of that, it is also important to devise adaptive toolkit management mechanisms where emerging multimodal interfaces can be easily incorporated. Such advancements could enhance the framework's capacity for complex reasoning while ensuring sustained relevance in dynamic information ecosystems.


\newpage
\bibliographystyle{ACM-Reference-Format}
\bibliography{sample-base}

\end{document}